\documentclass[conference]{IEEEtran}
\IEEEoverridecommandlockouts
\usepackage{cite}
\usepackage{amsmath,amssymb,amsfonts}
\usepackage{algorithmic}
\usepackage{graphicx}
\usepackage{textcomp}
\usepackage{xcolor}
\usepackage{comment}
\usepackage{booktabs}
\usepackage{url}
\usepackage{enumitem}

\usepackage{tikz,graphics,color,fullpage,float,epsf,caption,subcaption}

\def\BibTeX{{\rm B\kern-.05em{\sc i\kern-.025em b}\kern-.08em
    T\kern-.1667em\lower.7ex\hbox{E}\kern-.125emX}}
\begin{document}

\title{SpiroActive: Active Learning for Efficient Data Acquisition for Spirometry}

% \begin{comment}

\author{\IEEEauthorblockN{Ankita Kumari Jain}
\IEEEauthorblockA{
% \textit{dept. name of organization (of Aff.)} \\
IIT Gandhinagar, India \\
ankitajains@iitgn.ac.in}
\and
\IEEEauthorblockN{Nitish Sharma}
\IEEEauthorblockA{
% \textit{dept. name of organization (of Aff.)} \\
Independent Researcher, India \\
nitishsharma1295@gmail.com}
\and
\IEEEauthorblockN{Madhav Kanda}
\IEEEauthorblockA{
% \textit{dept. name of organization (of Aff.)} \\
IIT Gandhinagar, India \\
madhav.kanda@alumni.iitgn.ac.in}
\and
\IEEEauthorblockN{Nipun Batra}
\IEEEauthorblockA{
% \textit{dept. name of organization (of Aff.)} \\
IIT Gandhinagar, India \\
nipun.batra@iitgn.ac.in}
}
% \end{comment}

\maketitle

\begin{abstract}
Respiratory illnesses are a significant global health burden. %\nb{end line here and then delete the remaining part of the line.. till 2019.}%, 
Respiratory illnesses, primarily Chronic obstructive pulmonary disease (COPD), is the seventh leading cause of poor health worldwide and the third leading cause of death worldwide, causing 3.23 million deaths in 2019, necessitating early identification and diagnosis for effective mitigation. Among the diagnostic tools employed, spirometry plays a crucial role in detecting respiratory abnormalities. However, conventional clinical spirometry methods often entail considerable costs and practical limitations like the need for specialized equipment, trained personnel, and a dedicated clinical setting, making them less accessible. To address these challenges, wearable spirometry technologies have emerged as promising alternatives, offering accurate, cost-effective, and convenient solutions. The development of machine learning models for wearable spirometry heavily relies on the availability of high-quality ground truth spirometry data, which is a laborious and expensive endeavor. In this research, we propose using active learning, a sub-field of machine learning, to mitigate the challenges associated with data collection and labeling. By strategically selecting samples from the ground truth spirometer, we can mitigate the need for resource-intensive data collection. We present evidence that models trained on small subsets obtained through active learning achieve comparable/better results than models trained on the complete dataset.
\end{abstract}


\section{Introduction}
\noindent Respiratory illnesses pose a significant health challenge worldwide. Every year around 262 million people suffer from asthma, and 3.23 million deaths are attributed to Chronic Obstructive pulmonary disease (COPD)~\cite{Schluger2014LungDI} 
% \ra{Use ~ before the cite command like I have done here for all the citations}. 
With the emergence of a global pandemic like COVID-19, monitoring lung health has become an essential part of human survival~\cite{Achkar2022-bd}. Early detection and diagnosis of respiratory illnesses can mitigate their impact and reduce the risk~\cite{PMID:19716965,Van_Schayck2003-zp}.

\noindent Studies have indicated that the diagnosis of lung-related conditions is often missed or delayed until the health conditions have significantly worsened and clear signs of poor lung health are evident. One contributing factor to this issue is the relatively high cost of medical tests, which typically costs \$240\footnote{\label{fn:Vernier}Source: \url{https://www.vernier.com/product/spirometer/}}. Consequently, the availability of affordable and user-friendly equipment for diagnosing lung health is limited and poses a significant obstacle to timely diagnosis.~\cite{Brakema2019-hk}. 

\begin{table*}[h!]
\noindent
\resizebox{\textwidth}{!}{%
\begin{tabular}{@{}lrl@{}}
\toprule  
Dataset name & $\#$Participants & Type of Data\\
\midrule
SpiroSmart~\cite{Larson:2012:SUM:2370216.2370261} & 52 & Exhalation and Spirometer data \\
SpiroMask~\cite{10.1145/3570167} & 48 & Audio and Spirometer data \\
SpiroCall~\cite{goel2016spirocall} & 50 & Audio and Spirometer data \\
GRABMyo~\cite{Jiang2022-ii} & 43 & EMG data \\
MobSpiro~\cite{zubaydi2017mobspiro} & 25 & Audio and Spirometry data \\
Auditory evoked potential EEG-Biometric dataset~\cite{Abo_Alzahab2021-zz} & 20 & EEG recordings\\ 
Cough Dataset~\cite{Larson:2011:APP:2030112.2030163} & 17 & Audio data \\
BodyBeat~\cite{10.1145/2594368.2594386} & 14 & Body sounds like eating, breathing, speech etc. \\
DeepBreath~\cite{10.1145/3214289} & 13 & Sleep data \\
Wearable exam stress dataset~\cite{9744065} & 10 & electrodermal activity, skin surface temperature \\
 Abdominal and Direct Fetal ECG Database~\cite{Jezewski2012-kt} & 5 & FECG data\\ 

 \midrule\bottomrule 
 \end{tabular}%
 }
 \caption{Comparison of dataset size across various physiological signal based studies. The general trend across the table showcases the unavailability of large size datasets in the domain.}
 \label{demo-table}
 \end{table*}

\noindent Spirometry is a non-invasive pulmonary function test widely used to diagnose respiratory problems, and assess lung function~\cite{wiki:Spirometry}. Typically conducted in a clinical setting by trained clinician, the test requires certain preparatory measures such as avoiding smoking, consuming heavy meals, or engaging in tedious physical activity, as these factors can affect the accuracy of the results~\cite{clevelandclinic}. During the test, the patient exhales forcefully into a spirometer. The spirometer measures 23 different lung function parameters, such as forced vital capacity (FVC), peak expiratory flow (PEF) and forced expiratory volume in one second (FEV1), providing insights into respiratory health. 

\noindent Although spirometry helps assess the severity of lung diseases, performing clinical spirometry is expensive.
Mobile and wearable spirometry~\cite{gupta2011mobilespiro,Larson:2011:APP:2030112.2030163,Larson:2012:SUM:2370216.2370261} pose as affordable alternatives to clinical spirometry. These are cost-effective and have higher portability, enhancing accessibility. However, these approaches also have their drawbacks as accuracy is affected by factors like positioning of the device, the user's lip posture, and external environmental conditions~\cite{10.1145/3570167,goel2016extending} 

\noindent Verifying these techniques requires ground truth spirometry data, which is difficult to obtain due to logistical challenges in collecting consistent and reliable measurements across diverse environments and is hugely impacted by participants' cooperation and adherence to standardized testing procedures.
Thus datasets used to test out mobile spirometry approaches like SpiroSmart~\cite{Larson:2012:SUM:2370216.2370261} and MobSpiro~\cite{zubaydi2017mobspiro} are limited in participant size. The same trend is observed across majority of the physiological datasets. As seen in table~\ref{demo-table}, most prior studies related to physiological signals have been working with small amounts of data. As per~\cite{10.1145/3570167}, collecting a spirometry sample takes about 45 minutes to an hour and relies heavily on the patient's ability to follow clinical instructions, which can be challenging, especially for young and elderly patients. 

\noindent The core premise of active learning is that a machine learning system can improve its accuracy if it is allowed to select the data it learns from. In this research, we propose an active learning approach using a publicly available spirometry dataset~\cite{10.1145/3570167}, which contains estimated values for FEV1, FVC, and PEF obtained from a wearable spirometry study. We aim to investigate the benefits of active learning by selecting the most informative samples that lead to a reduction in errors from previous baselines. To evaluate our approach, we explore different learning strategies like Single Task and Multi-Task Active Learning and Single Output and Multi-Output (which are further discussed in the upcoming sections). 
% \nb{recgheck this entire para; also include the main results from our study .. x\% samples..main method -- multi-task AL..weighing function...}
\noindent The American Thoracic Society (ATS) allows up to~7\% error rate between spirometry maneuvers
~\cite{10.1145/3570167}. Our proposed methods consistently outperform this with error rates of 4.96\% for FEV1 and 4.45\% for FVC, using Leave-One-Out cross-validation. We achieve these errors using active learning with $\sim$30\% of the total dataset in the individual task learning setting (single-task setting). 

\noindent Since collecting participant data specific to individual tasks [FVC, FEV1] is impractical, we propose leveraging the joint distribution across output parameters to select common points across both parameters. By applying various weighting schemes and utilizing models that inherently support multi-output settings (such as Random Forest and Decision Trees), we can strategically leverage the correlation between outputs and improve performance.

\noindent To summarize, the main contributions of this research work are: 
\begin{itemize}[left=0pt]
    \item Active Learning for Spirometry: We propose using active learning for spirometry to reduce data acquisition costs while ensuring same or better performance
    using a subset of data ($\sim$24 participants).

    \item Multi-Task Approaches: Models and weighing schemes that can use the correlation between the output parameters and jointly optimize over them to reduce errors across tasks.
    \item Reproducibility: We publicly release our codebase and believe that our work is fully reproducible. All the generated tables and graphs have corresponding scripts to reproduce all the results \footnote{\label{fn:gitrepo}Source: \url{https://github.com/percom-anonymous/Spiroactive}}
\end{itemize}

\section{Background}
\noindent This section discusses about the pertinent background for clinical spirometry, the recent advances in mobile and wearable spirometry and briefly introduces the concept of active learning.

\subsection{Clinical Spirometry} \label{spirometry}
\begin{figure}
 \centering
  \includegraphics[scale = 0.7]{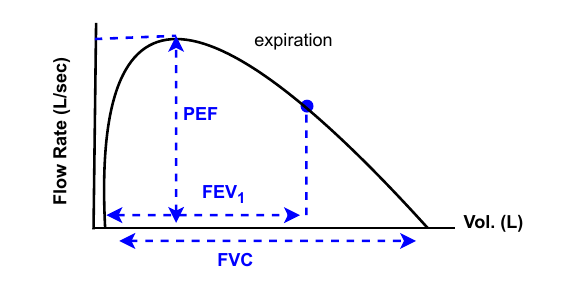}
        \caption{Graphical Representation of the flow-volume curve in spirometry showing the three commonly used lung health indices : Peak Expiratory Flow (PEF), Forced Expiratory Volume in 1 second (FEV1) and Forced Vital Capacity (FVC).}
    \label{fig:flow_curve}
\end{figure}

\noindent Spirometry is a pulmonary test widely used to assess breathing patterns and detect lung ailments like asthma, cystic fibrosis, and COPD~\cite{wiki:Spirometry}. During a spirometry maneuver a patient breathes into a spirometer mouthpiece which measures the amount and speed of air exhaled. Spirometers processes the collected flow-volume curves (Figure~\ref{fig:flow_curve}) to extract lung paramters. The American Thoracic Society (ATS) discusses different lung health parameters that are extracted from spirometry. 
The three majorly used parameters are: 
\begin{itemize} [left=0pt]
    \item FVC - Forced vital capacity is the total air volume exhaled (Litres).The plot of the expired volume illustrates the FVC as the peak of the graph, representing the maximum expiration volume. 
    \item FEV1 - Forced expiratory volume in 1 second is the exhaled air volume in the first second of exhalation (Litres).
    \item PEF - Peak expiratory flow is the maximum airflow velocity in exhalation. (L/s). The plot of the flow rate illustrates this at the peak value in the expiration stage.
\end{itemize}

\subsection{Mobile and Wearable Spirometry}
\noindent Previous efforts have investigated alternative systems ~\cite{Larson:2012:SUM:2370216.2370261,zubaydi2017mobspiro} due to challenges in clinical Spirometry such as: i) Difficulty in ensuring patient cooperation, especially with young children and elderly patients; ii) Spirometry instrumentation is expensive and requires time-consuming efforts for data collection. To address these issues, cheaper and more accessible alternatives such as wearable and smartphone spirometry have been studied in the past. ~\cite{agu2013smartphone,gupta2011mobilespiro,zubaydi2017mobspiro,viswanath2018spiroconfidence}. Recent efforts have also been made to incorporate wearables (such as masks) with audio and pressure sensors to extract vital lung parameters~\cite{piuzzi2020wearable,enokibori2013spirovest,fischer2021masquare,costa2018wearable,10.1145/3570167,zhou2020accurate}. They provide a more secure and enclosed environment to capture audio data, that is used as a proxy for the airflow, to estimate lung health indices such as FEV1, PEF and FVC.
\noindent Despite promising results, studies in this area usually consists of small participant groups (typically $<$70 as shown in Table \ref{demo-table}) owing to difficulty in collecting ground truth spirometry data. The efficiency of active learning in low data settings coupled with the convenience of hand held spirometry motivated us towards studying active learning for mobile spirometry.

\subsection{Active Learning}\label{ActiveL}
\begin{figure}
 \includegraphics[scale=0.55]{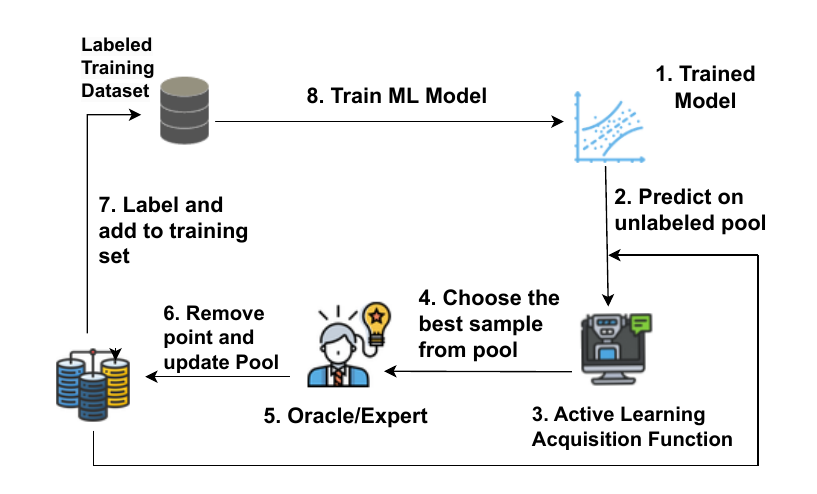}
 \caption{Active Learning Loop that showcases the flow of data across the whole pipeline}
 \vspace{-10pt}
\label{fig: AL_loop}
\end{figure}
\noindent One of the main reasons behind the success of machine learning in the recent past has been the availability of large datasets. Collecting large quantities of \textbf{unlabeled} data in medical and imaging settings is often easy. However, obtaining the \textbf{labels} for such datasets is usually time-consuming and expensive.

\noindent Active Learning is a sub-field of machine learning that aims to reduce the labeling or annotation cost by strategically labeling samples from the large unlabeled set. The active learning scenarios can be broadly divided into three types~\cite{settles2009active}: i) Membership Query Synthesis; ii) Stream-Based Selective Sampling and iii) Pool-Based Sampling. We direct the reader to a comprehensive active learning survey~\cite{settles2009active} and focus on the setting most pertinent for our application: pool-based sampling. 

\noindent Pool-based sampling is used in many real-world problems like text classification~\cite{tong2001support}, information extraction~\cite{wu2005semi}, speech recognition~\cite{riccardi2005active}, and cancer diagnosis~\cite{liu2004active}. Pool-based sampling has also been widely used to reduce human annotations in the activity recognition task.~\cite{al2,al3} \noindent Broadly, pool-based sampling is based on labeling samples from a large pool of unlabeled data instances via heuristics called acquisition functions. We now present the active learning workflow for pool-based sampling, as illustrated in Figure~\ref{fig: AL_loop}.
\begin{enumerate}[left=0pt]
    \item      
    % \ra{can you label this figure? As I read, I do not know which part of the figure should I focus on. Handhold the reader to the parts of the figure and make it easy for her/him to comprehend.},\ns{Done please review the new figure}   
    The active learning loop consists of a machine learning model initially trained on the labeled dataset (usually small initially).
    \item The trained model is then used to make predictions on the large pool set. 
    \item We use acquisition function, $A(x)$, to calculate scores for the pool set. In this context, the variable $x$ denotes the data points within the pool set.

    \item We choose the next point to be queried as the one which maximizes the acquisition function.
    $\Bar{\boldsymbol{x}}_*=\operatorname{argmax}_{x_* \in x_{\text {pool }}} A(\boldsymbol{x}_*)$     
    
    \item We obtain the true label/ground truth (${\boldsymbol{y}}_*)$ for the chosen pool point $\Bar{\boldsymbol{x}}_*$ 
    
    \item We add the point ($\Bar{\boldsymbol{x}}_*, {\boldsymbol{y}}_*$) to the training dataset and correspondingly remove $\Bar{\boldsymbol{x}}_*$ from the pool dataset.
    \item Continue with Step 1 iteratively until the allocated budget, representing the maximum number of queries or labeled samples, is depleted, or until the predefined stopping condition is met within the active learning framework.
   
\end{enumerate}

\noindent For $x_*$, we can look at uncertainty in output $y_*$ predicted by the current model and use the variance as the acquisition function: $A\left(\boldsymbol{x}_*\right)=\operatorname{var}\left(\boldsymbol{y_*}\right)$. Generally, two kinds of methods are used in the active learning literature for regression: 

% \ra{acquisition function should have been mentioned earlier as I pointed out before. I think the points above will be easier to read if you briefly introduce the reader to the variables used} 

\begin{itemize}[left=0pt]
    \item \textbf{Uncertainty Sampling:} Here we can use machine learning models that inherently provide uncertainty estimates and use this uncertainty as our acquisition. 
    \item \textbf{Query by committee (QBC):} Here we can create a committee of multiple machine learning models and use the variance in prediction across this committee as our acquisition.
\end{itemize}

\section{Problem Statement}

\noindent Mobile/wearable spirometry systems collect audio data while users perform forced breathing maneuvers. The collected raw audio data undergoes normalization and Hilbert transformation to generate its envelope. Various features are extracted from the audio as per ~\cite{Larson:2012:SUM:2370216.2370261} and the ground truth lung health indices are collected using a forced breathing spirometer maneuver 

\noindent Collecting wearable audio data is significantly easier as compared to acquiring the ground truth spirometer data, which requires expert supervision and clinic visit. Thus, we pose our active learning problem as follows: 
\begin{itemize}[left=0pt]
    \item We assume a large number of samples (pool data) of audio data and corresponding features collected from mobile/wearable spirometry setups.
    \item  We assume  only a small fraction of samples (training data) to be associated with ground truth data containing the two lung health indices.
    \item Our goal is to ``strategically'' choose participants from the pool set and obtain their true labels.
    \item The role of the oracle in our active learning setting is akin to visiting the lab where the ground truth spirometry can be performed.
\end{itemize} 

\noindent The overall vision is to get ``good'' performance using a small number of points sampled from the pool. Prior literature considers a 7\% mean absolute percentage error between our estimated lung health indices and ground truth to be a ``good'' performance per the ATS guidelines. ~\cite{Larson:2012:SUM:2370216.2370261,10.1145/3570167,PMID:31613151}\\

\section{Methodology}

\subsection{Learning Paradigms} \label{task-setting}

\noindent We now present our method, focusing on the primary lung health indices: FEV1, FVC.
Prior mobile/wearable spirometry literature has proposed distinct ``tuned'' features for each health index based on domain knowledge ~\cite{10.1145/3570167}. For example, FEV1 is the volume of air exhaled in the first second of a forced expiratory maneuver, so signals beyond 1s are not needed. In contrast, measuring FVC requires the complete signal. Given the different set of (partially overlapping) features, we consider estimating FEV1, and FVC, as two different tasks. For our experimentation, we use the Single Output setting where we train a different model for each individual task. Each trained model, though possibly of the same class has its own distinct parameters ($\theta$) and hyperparameters ($\phi$). Previous literature~\cite{10.1145/3570167, Larson:2012:SUM:2370216.2370261} has generally focused on this experimental setting. 

% \subsection{Multi-Output Modelling}
% \nb{Ok I am confused are we learning separate params per lung param as previously mentioned or learning same params/shared params as we mention in this section..}

% \noindent Multi-Output models are a set of machine learning models that can simultaneously predict multiple output variables. They need optimization methods to jointly optimize over multiple labels. These models can potentially offer the advantage of capturing the correlation between output values. This helps improve the overall performance and interpretability of the model. While Neural Networks are widely used for multiple output predictions, open source implementations for multi-output variants for many standard machine learning models are not readily available. In this study, we use multi-output decision trees (and thus multi-output Random Forests). In the single output settings, regression decision trees split along an input feature so as to maximize the reduction in mean squared error. 
% We modify the single-output decision tree algorithm by choosing splits that optimize the reduction in mean squared errors across the two outputs simultaneously. 
% We do not use multi-output multi-layer perceptrons (MLPs) due to reasons discussed in the upcoming sections. 

\subsection{Active Learning Settings} \label{AL-setting}

\noindent As discussed in section~\ref{ActiveL}, the active learning loop queries points that improve model performance. The querying for pool points can be done in two ways:
\begin{enumerate}
    \item Single-Task Active Learning where different points are queried for different tasks.
    \item Multi-Task Active Learning where the same point is queried for two tasks
    
\end{enumerate}
\textbf{Single-Task Active Learning} 

In this setting, the two tasks (FEV1 \& FVC) 
% \ra{which three?} 
are treated as two independent problems. The querying for each task, at each iteration, is done based on the given function:
\begin{center}
    \centering
$\Bar{\boldsymbol{x}}_*=\operatorname{argmax}_{x_* \in x_{\text {pool }}} A(\boldsymbol{x}_*)$
\end{center}

\noindent where $A\left(\boldsymbol{x}_*\right)=\operatorname{var}\left(y_*\right)$
This leads to difference in the order in which pool points are selected in each task. Single Task AL can be achieved by using the single output models as described in section~\ref{task-setting} .

\noindent \textbf{Multi-Task Active Learning}

While Single-Task AL tends to achieve the best results as the problems are independent of each other, it is often wasteful to acquire separate participant data for each task. This section discusses the importance of using a single approach to pick points that lead to error reduction across all the tasks. The MultiTask AL acquistion function using weighing can be stated as:
\begin{center}
    \centering
$A\left(\boldsymbol{x}_*\right)=\sum_{m=1}^{M}{\operatorname{var}\left(y_*^{m} \right)* w_m}$
\end{center}

\noindent where $M$ is the number of outputs (e.g. $M=2$ for our problem corresponding to FEV1, and FVC), $y_*^m$ is the output for task $m$ and $w_m$ is the weight associated with the $m^{th}$output. 
We use various weighing techniques to determine the weights $w_m$: 
\begin{enumerate}[left=0pt]
    \item Uniform weighted uncertainty: We uniformly weigh the uncertainties across all the tasks and pick the most uncertain data point from the pool. 
    \begin{center}
        \centering
        $w_m = \dfrac{1}{M}$
    \end{center}
    \item Maximum Weighted uncertainty: The weighing factor gives the highest weightage to the task with the highest uncertainty.
    \begin{center}
        \centering
        $w_m= \dfrac{\sum_{i=1}^{N}{var(y_{i*}^m)}}{\sum_{m=1}^{M}{\sum_{i=1}^{N}{var(y_{i*}^m)}}}$
    \end{center}

    where $N$ is the total number of samples in the pool, $M$  is the total number of outputs and $y_{i*}^m$ is the prediction for the $m^{th}$ task corresponding to the $x_{i*}$ point from the pool.
    \item Inversely weighted uncertainty: The task with the maximum uncertainty is given minimum weightage. 
     \begin{center}
        \centering
        $w_m= \left(\dfrac{\sum_{i=1}^{N}{\operatorname{var}(y_{i*}^m)}}{\sum_{m=1}^{M}{\sum_{i=1}^{N}{\operatorname{var}(y_{i*}^m)}}}\right)^{-1}$
    \end{center}
\end{enumerate}

The other Multi-task AL acquisition functions include: 
\begin{itemize}[left=0pt]
    \item Rank Sampling: Each data point is ranked by uncertainty across models. The ranks are summed, and the point with the highest uncertainty/lowest rank sum is selected for active learning.
    \begin{center}
        \centering
    $A\left(\boldsymbol{x}_*\right)=\operatorname{argmax}\left(\sum_{m=1}^{M}{\operatorname{var}\left(y_*^{m} \right)}\right)$
    \end{center}
    
    \item Round Robin Sampling: In this technique, we query the most uncertain points across each task in a round robin fashion, i.e., in the first iteration, the point is queried according to the first task and so on. The round-robin sampling technique ensures that each task is given an equal number of samples, promoting fairness and reducing bias in the AL loop.
\end{itemize}

\section{Evaluation}

\subsection{Dataset}
\noindent For our experimentation, we have collaborated with SpiroMask authors~\cite{10.1145/3570167} and collected 10 more samples to expand our dataset. Out of the total 58 samples (including 10 new sample), we remove 7 samples based on the Flow-Volume envelope and the back-extrapolated volume (BEV) ~\cite{PMID:31613151}.

\noindent Out of \textbf{51 participants} (Table~\ref{dataset}), \textbf{10} suffer from various lung ailments (asthma, obstructive lung ailments, wheezing, COVID infections, and other restrictive ailments). The study consists of \textbf{16 female} participants, and the age group of people varies from \textbf{21 to 68 years} ~\cite{10.1145/3570167}.
 
\noindent To generate the feature set for active learning, we pass the audio signal through a Butterworth bandpass filter with cut-off frequencies of \textbf{3000 and 5000 Hz} that conserve all the tidal breathing information.

\noindent The filtered audio is used to generate acoustic and temporal features like Spectrogram features (STFT), Mel Spectrogram, Mel-frequency energy (MFE), Mel-frequency cepstral coefficients (MFCC), and spectral features. We then apply Sequential Forward Selection to jointly optimize our dataset for both the output variables ~\cite{Larson:2012:SUM:2370216.2370261},~\cite{10.1145/3570167},~\cite{Larson:2011:APP:2030112.2030163}. Although PEF is commonly used for lung health prediction, discussions with the dataset authors revealed that a small number of samples were clamped partly impacting the signal and peak required for calculating PEF. However, it does not affect the features needed for FVC and FEV1. Hence, PEF is not considered for future experimentation.

 \begin{table}[h!]
 \centering
 \begin{tabular}{@{}lr@{}}\toprule 
 \midrule
 Total participants & 51\\ 
  Participants with Lung ailments & 10\\ 
 Female participants & 16\\ 
 Age (yrs) (mean, range) & 28.05 (21 - 68)\\ 
 Height (cm) (mean, range) & 167.39 (142.4 - 182.9)\\ 
 \midrule\bottomrule 
 \end{tabular} 
 \caption{Participant Demographics}  \label{dataset}
 % \zb{Tables can be made with booktabs package using toprule and bottomrule}
 \end{table}

 \subsection{Metrics}
 \noindent We use the mean absolute percentage error to compare performance across all the participants for FVC, and FEV1 tasks. 
\begin{center}
    $ MAPE=\left(\frac{1}{N}\sum _{i=1}^{N}\left(  |\frac{y- \hat{y}}{y} | \right)\right) *100 $
\end{center}
\noindent where $y$ is the true value, $\hat{y}$ is the predicted value and $N$ is the total number of test points. 
\noindent The guidelines on standardized spirometry~\cite{PMID:31613151} and later studies~\cite{Larson:2012:SUM:2370216.2370261, doi:10.1080/09291010903016135} describe the error between two spirometry efforts in percentages. Hence we choose MAPE scores as the measurement method. 

\subsection{Experiments}
\noindent In this section, we discuss the experiments and their results to study the efficacy of active learning for spirometry. For our experiments, we use a committee consisting of Random Forest Regressor, Decision Tree Regressor, and Gradient Boosting Regressor, each initialised with different random seeds. This committee is formed based on an analysis of LOOCV results (details provided in the online appendix \footnotemark[\value{footnote}]).
% \subsubsection{\textbf{Experiment 1: Do we benefit by using shared data and model representations?}} \label{multi-task}

% \noindent In our first experiment , we explore the advantages of utilizing shared representations of data and compare the two ML model task settings discussed earlier. 
% In section~\ref{spirometry}, we discuss the flow-volume curve used to derive the three lung indices. While each task requires a separate feature representation for optimum performance, it is not feasible to acquire participant data specific to a single task. To address this, we use leave-one-out cross validation, and Committee and Random Forest models for our experimentation. 
% While employing a joint feature set for both tasks could potentially reduce data acquisition costs, it is counterproductive to the model's performance. This is because not all features in the superset are necessary for accurate predictions and including irrelevant features may introduce noise into the data. 
% The committee, in single output setting, for the different tasks is Random Forest Regressors, Decision Tree Regressor and Gradient Boosting Regressor with different initialisations.

\subsubsection{\textbf{Experiment 1: How do we choose the train-pool-test split for evaluation in the Active Learning settings?}}

% \nb{what methods are you using for modelling? why? refer to Adhiokary et al..}

\begin{figure}
    \centering
    \includegraphics[scale=0.6]{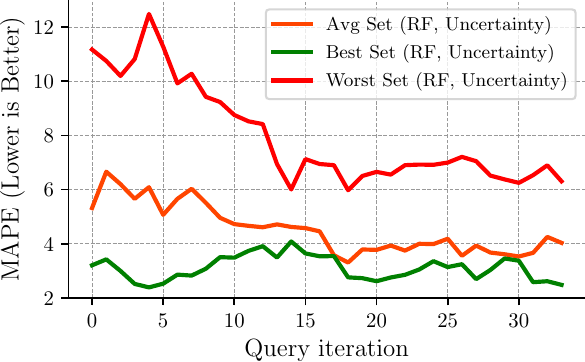}
    \caption{MAPE scores for different train-test splits for the FVC task. The figure shows the importance of choosing a representative yet diverse split, especially in low data settings.}
    \label{fig:train_test_split}
    \vspace{-10pt}
\end{figure}

\noindent Real-life scenarios, especially medical settings including our spirometry dataset, are low-resource, low-budget problems and do not have a large labeled pool. This section examines how different splits affect active learning outcomes. We divide the data into train, test and pool sets, arguing that the choice of initial training set determines initial model quality and uncertainties evaluated on the pool points further influence this. 

\noindent There are two extreme scenarios when selecting the train set: highly representative (best set) or highly diverse (worst set) compared to the test data. In practice, we would end up with an average set that has equal notions of diversity and representation. 

\noindent Finding the different best, average and worst sets would require us to enumerate over all possible options of train-test splits which is prohibitively expensive. Instead, we perform active learning multiple times using Random Forest ensembles on various train-test splits, with \textbf{8 training} and \textbf{10 test} points in each split. Figure~\ref{fig:train_test_split} showcases the MAPE scores for three different train-pool-test splits. The best set achieves scores as low as \textbf{4\% without any pool data}, indicating a representative distribution. However, using the best set negates the need for active learning, while the worst set leads to high initial errors and does not have same data representation in the train-test split. Though this split also beats the 7\% ATS error requirements and follows the active learning trend of error reduction, it is still far off from the other splits in terms of the lowest errors it can achieve. 

\noindent For further experimentation, we use the average train-test sets to ensure representation while retaining the randomness in data. Thus the choice of train-pool-test for active learning, especially in low data settings is very important. We propose a heuristic/simulation to create these splits and observe the relevance of this problem through the error performance.

\subsubsection{\textbf{Experiment 2: What is the lowerbound on the Mean Absolute Percentage Error for Active Learning?}}

In this section, we motivate the need for active learning in clinical settings like the SpiroMask~\cite{10.1145/3570167}. We establish an oracle baseline that uses brute-force approach to greedily find the (best parameters, pool point) pair in each iteration that maximally increase the model performance on the average set selected in the previous experiment. The random sampling strategy randomly queries a pool point at each iteration, and the experiment is repeated \textbf{20 times} to ensure pure randomness. 

\noindent We average out the predictions across all the tasks to plot the mean and obtain the 2$\sigma$ value to plot the 95\% confidence interval. Since the oracle greedily picks the points that minimise the error, we end up exhausting the best set of points in few iterations and are left with noisy points that lead to an increase in the error rates. 

\begin{figure}
    \centering
    \includegraphics[scale=0.6]{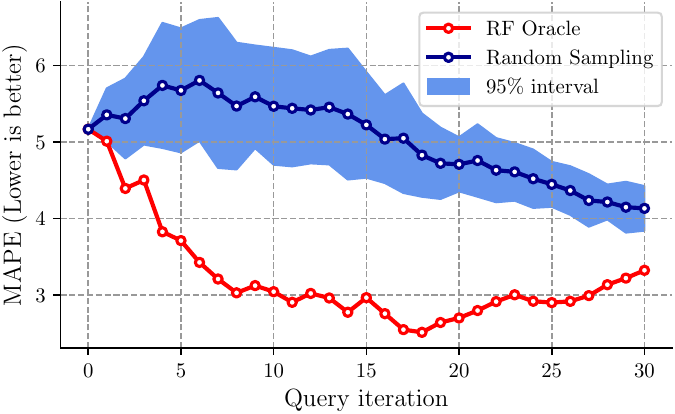}
    \caption{Oracle vs Random Sampling for FVC Dataset. The oracle beats random sampling by a huge gap indicating the need for better acquisition strategies and intelligent models}
    \label{fig:oracle_vs_random}
    \vspace{-15pt}
\end{figure}

\noindent Figure~\ref{fig:oracle_vs_random} showcases a significant gap between the random acquistion and oracle baselines. The oracle achieves errors as low as \textbf{2.39\% with 57\% data points}, while the random sampling strategy does not cross the 4\% error line even with the entire dataset. This difference in error indicates that intelligent querying strategy can help bridge the gap between the oracle and random sampling giving us an efficient sampling technique.

\subsubsection{\textbf{Experiment 3: What is the performance of Active Learning in Single Task setting?}}

\noindent In Single task setting, we train separate models for FVC and FEV1 using two querying strategies - \textbf{Query-Based Committee} and \textbf{Uncertainty-based sampling} to acquire points that help achieve lower errors, thus reducing computation and labeling costs. We propose a committee of regressors and use Query by committee as a strategy to query points across the various regressors in the committee. 

\begin{figure*}
    \centering
    \begin{subfigure}{.5\textwidth}
    \centering
    \includegraphics[width=0.7\textwidth]{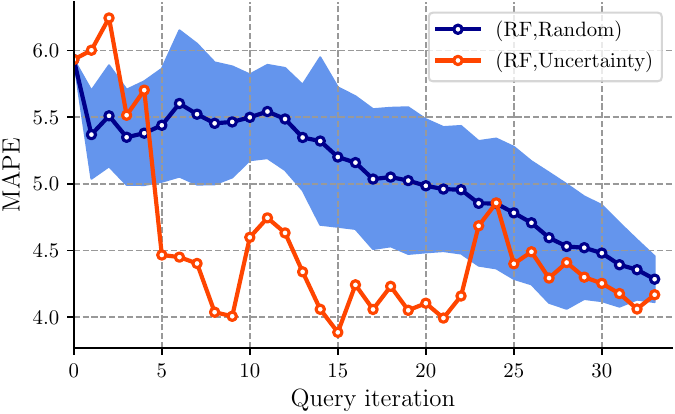}
    \caption{FEV1}
    \end{subfigure}%
    \begin{subfigure}{.5\textwidth}
    \centering
        \includegraphics[width=0.7\textwidth]{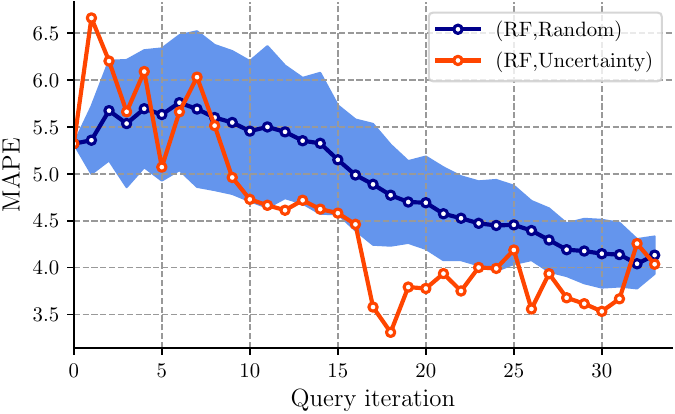}
        \caption{FVC}
    \end{subfigure}
    \caption{The curves for active learning using Random Forest and Standard Deviation as query strategy vs Random Sampling. It can be seen that we conveniently beat random sampling by a large value using lesser points.}
    \label{fig:single_op_rf}
\end{figure*}

\noindent Figure~\ref{fig:single_op_rf} shows the performance of Random Forest regressors, where the uncertainty is calculated from standard deviations across trees in the ensemble. We observe that our proposed model beats the random baselines by a significant margin. 
For FEV1, the AL-based RF model achieves errors as low as \textbf{4\% with just 39\%} of the entire training dataset , while for FVC, we reach errors as low as \textbf{3.3\% with 63\%} of the dataset. In case of FVC , performance similar to the full dataset is achieved with just \textbf{$\sim$61\%} of the points.

\begin{figure*}
    \centering
    \begin{subfigure}{.5\textwidth}
    \centering
    \includegraphics[width=0.7\textwidth]{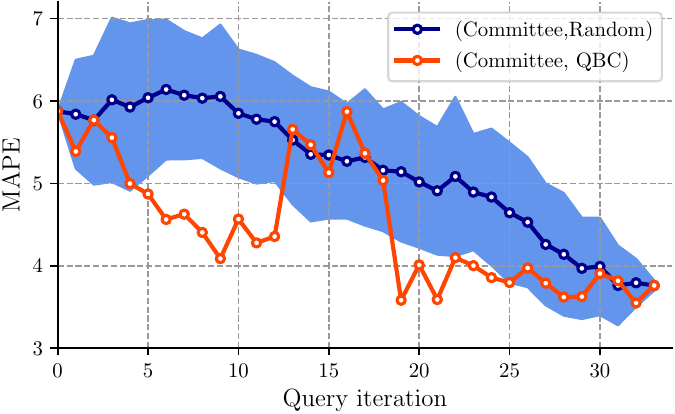}
    \caption{FEV1}
    \end{subfigure}%
    \begin{subfigure}{.5\textwidth}
    \centering
        \includegraphics[width=0.7\textwidth]{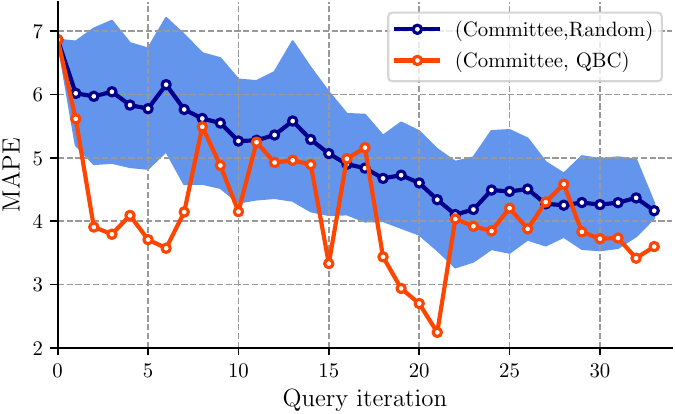}
        \caption{FVC}
    \end{subfigure}
    \caption{The curves for active learning using Committee Regressor vs Random Sampling. It can be seen that we beat random baselines and achieve lower errors for FVC task.}
    \label{fig:single_op_qbc}
\end{figure*}

\noindent Figure~\ref{fig:single_op_qbc} shows results using committee regressors, which also consistently outperform random baselines. We reach errors as loss as \textbf{3.5\%} for FEV1 and \textbf{2.3\%} for FVC with just 66\% and 71\% of the entire training dataset. Notably, we achieve performance comparable to the full dataset with just 27 points for FEV1 and 23 for FVC. This highlights the value of leveraging model uncertainty to reduce labeling costs. 

\noindent Although both techniques beat the random baseline, the committee regressors achieve the lowest errors. However these errors are achieved later in the active learning cycle as compared to the RF models, indicating a trade-off between the number of AL iterations and the errors achieved. We observe that the total training errors obtained from our proposed acquisitions and random sampling do not intersect at the last iteration. This occurs because Random Forest models use bootstrapping for training data, meaning that each tree is trained on a different subset of data based on the sequence of training points, resulting in varied predictions.

\subsubsection{\textbf{Experiment 4: How practical is Single Task Active Learning?}}

\begin{figure}
        \centering
        \includegraphics[scale=0.8]{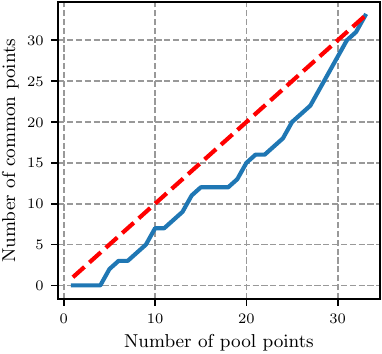}
        \captionof{figure}{ Intersection of the pool points selected in each iteration. As expected, each task picks a different pool point and thus the intersection up-to 5 iterations remains zero.}
        \label{pool_intersect}
        \vspace{-20pt}
\end{figure}

\begin{figure*}
\centering
        \includegraphics[width=0.7\textwidth]{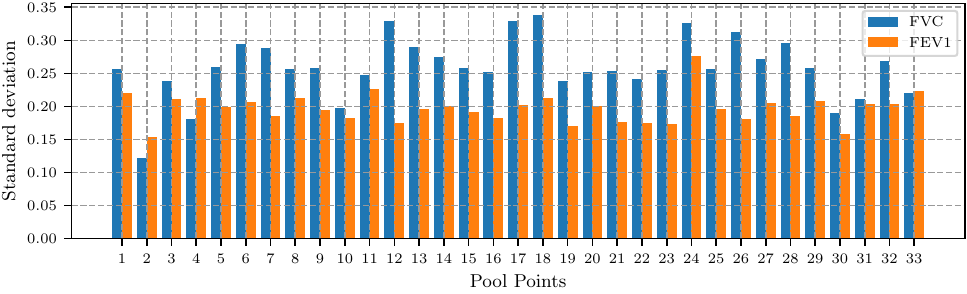}
    \caption{A comparison of uncertainties obtained after the first iteration for pool points for different tasks.It can be seen that both the task pick 2 different points as the most uncertain. Thus there is a need to perform multi-task active learning.}
    \label{fig:comparison_stso_uncertanity_pool_points}
\end{figure*}

% \nb{in general for all our figures, we need to expand the writeup to ensure that the main message and explanation is coming clearly.}
\noindent While Single task active learning gives the best performance, it is not feasible to acquire individual participant data for each task. As shown in figure~\ref{pool_intersect}, there are no common points across all the tasks until the 5$^{th}$ iteration, indicating that each task performs independently. Figure~\ref{fig:comparison_stso_uncertanity_pool_points} further illustrates that different tasks prioritize different points in the pool dataset. It is rarely observed that a single point achieves highest uncertainty across all the tasks. 

\noindent To address this, we propose the usage of Multi Task Active learning, using different weighing schemes and models that leverage the relationship between both the tasks. By combining uncertainties from both tasks, an optimal query point can be selected in each iteration. We achieve this by using a combination of single-output models and different weighing schemes.

\begin{figure*}
    \centering
    \begin{subfigure}{.33\textwidth}
     \centering
        \includegraphics[width=\textwidth]{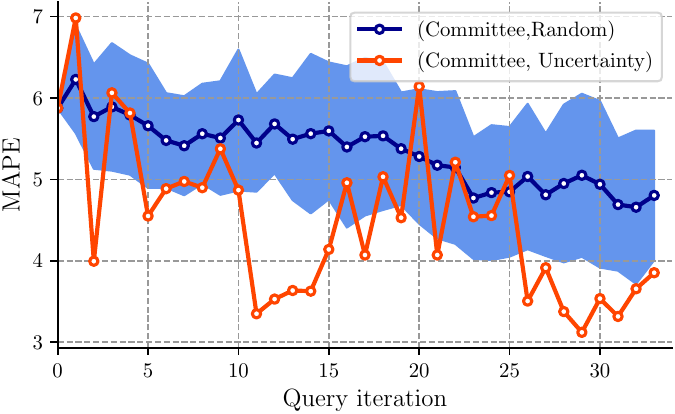}
        \caption{Weighted Query}
        \label{FEV_std}
    \end{subfigure}%
    \begin{subfigure}{.33\textwidth}
    \centering
    \includegraphics[width=\textwidth]{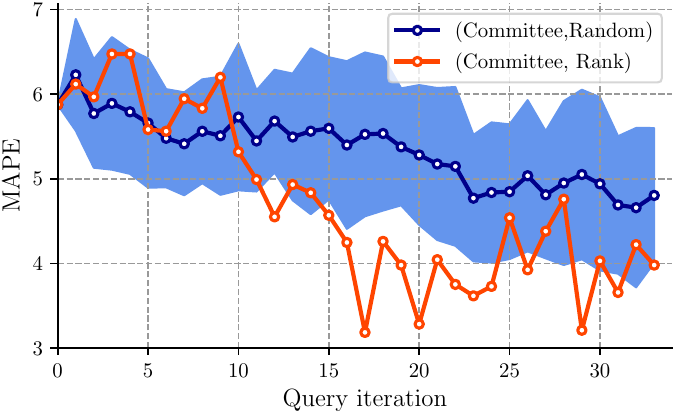}
    \caption{Rank Based}   
    \label{FEV_rank}
    \end{subfigure}%
     \begin{subfigure}{.33\textwidth}
    \centering
    \includegraphics[width=\textwidth]{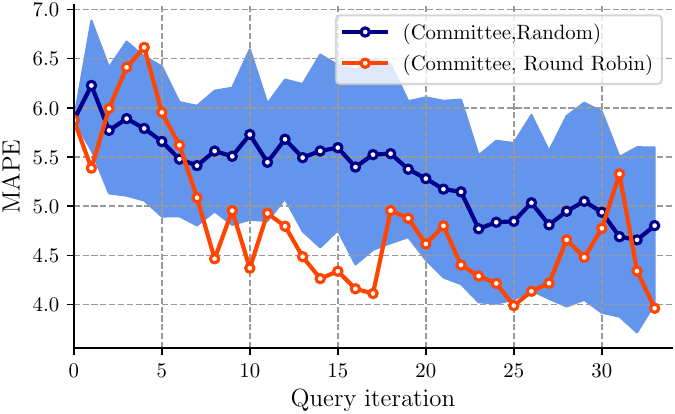}
    \caption{Round Robin}   
    \label{FEV_round}
    \end{subfigure}
    \caption{Single Output Querying using Committee - FEV1}
    \label{FEV1}
\end{figure*}

\begin{figure*}
    \centering
    \begin{subfigure}{.33\textwidth}
    \centering
    \includegraphics[width=\textwidth]{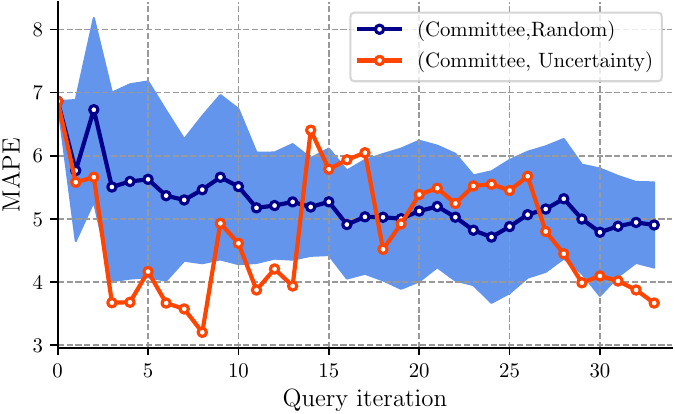}
    \caption{Weighted Query}
    \label{FVC_std}    
    \end{subfigure}%
    \begin{subfigure}{.33\textwidth}
    \centering
    \includegraphics[width=\textwidth]{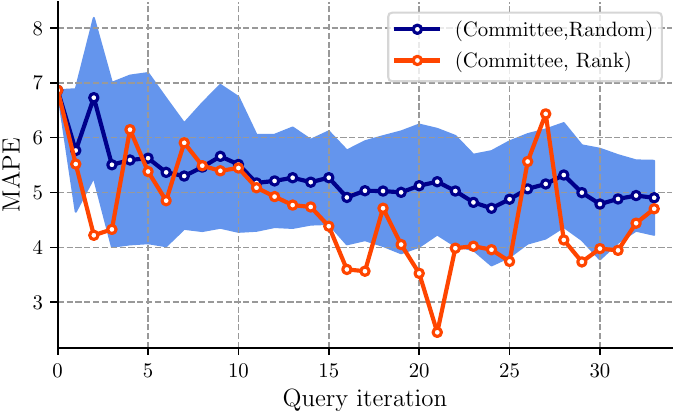}
    \caption{Rank Based}
    \label{FVC_rank}
    \end{subfigure}
     \begin{subfigure}{.33\textwidth}
    \centering
    \includegraphics[width=\textwidth]{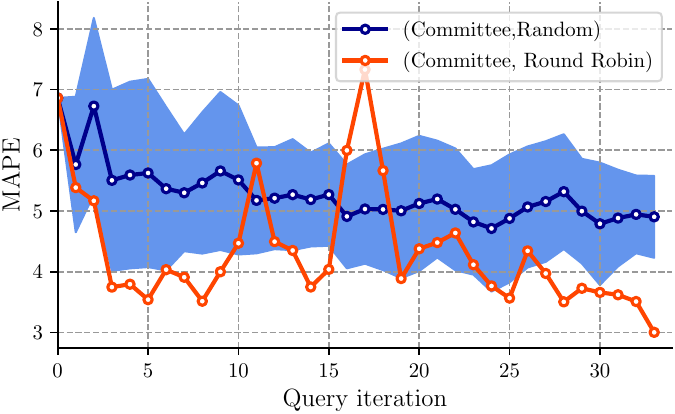}
    \caption{Round Robin}
    \label{FVC_round}
    \end{subfigure}
    \caption{Single Output Querying using Committee - FVC}
    \label{FVC}
    \vspace{-12pt}
\end{figure*}

% \begin{figure*}
%     \centering
%     \begin{subfigure}{.5\textwidth}
%      \centering
%         \includegraphics[width=0.8\textwidth]{images/FEV1_std_rf.pdf}
%         \caption{FEV1}
%     \end{subfigure}%
%     \begin{subfigure}{.5\textwidth}
%     \centering
%     \includegraphics[width=0.8\textwidth]{images/FVC_std_rf.pdf}
%     \caption{FVC}   
%     \end{subfigure}
%     \caption{Single Output Uncertainty based querying using Random Forest}
%     \label{fig:single_op_std_rf}
% \end{figure*}

% \begin{figure*}
%     \centering
%     \begin{subfigure}{.5\textwidth}
%     \centering
%     \includegraphics[width=0.8\textwidth]{images/FEV1_rank_rf.pdf}
%         \caption{FEV1}
%     \end{subfigure}%
%     \begin{subfigure}{.5\textwidth}
%     \centering
%     \includegraphics[width=0.8\textwidth]{images/FVC_rank_rf.pdf}
%     \caption{FVC}
%     \end{subfigure}
%     \caption{Single Output Rank based Querying using Random Forest}
%     \label{fig:single_op_rank_rf}
% \end{figure*}

% \begin{figure*}
%     \centering
%     \begin{subfigure}{.5\textwidth}
%      \centering
%         \includegraphics[width=0.8\textwidth]{images/FEV1_alternate_rf.pdf}
%         \caption{FEV1}
%     \end{subfigure}%
%     \begin{subfigure}{.5\textwidth}
%     \centering
%     \includegraphics[width=0.8\textwidth]{images/FVC_alternate_rf.pdf}
%     \caption{FVC}   
%     \end{subfigure}
%     \caption{Single Output Round Robin querying using Random Forest}
%     \label{fig:single_op_alternate_rf}
% \end{figure*}

\subsection{\textbf{Experiment 5: What is the performance of Multi Task Active
Learning?}}
\noindent Figures~\ref{FEV_std} and~\ref{FVC_std} show the results of weighted uncertainty-based querying, where different weights are assigned to the different task uncertainties. In each iteration, the pool point with the highest sum of uncertainties is selected for querying. This approach outperforms random sampling and achieves performance close to using the entire dataset, with a significantly smaller fraction of points. Using weighted querying by committee, we reduce errors to \textbf{3.3\%} on both FVC and FEV1 tasks with just \textbf{16 and 20} training points, respectively. The results from the Random Forest ensemble can be found in the online appendix.

\noindent Rank based sampling (\ref{FEV_rank},~\ref{FVC_rank}) also delivers errors as low as \textbf{3.1 \& 2.6\%} using the committee and \textbf{3.4 \& 4.0\%} using RF for FEV1 and FVC respectively. We also achieve performance closer to the entire dataset with a maximum of 28 training points.

\noindent Round Robin sampling (\ref{FEV_round},~\ref{FVC_round}) alternates the selection of points based on tasks, which can sometimes lower errors for one task while increasing them for another. This suggests that if we only used FVC or FEV1 for querying, we might have selected that point during the initial or final iterations. Despite this, Round Robin sampling still outperforms random sampling in the majority of iterations and achieves overall performance with a small fraction of points.

\section{Limitations and Future Work}

\noindent We now discuss some limitations of our present work and how we plan to address them in the future.
\begin{itemize}[left=0pt]
    \item We presently assume uniform labeling costs associated with each sample, but in reality, different samples may have varying costs. For example, older participants are harder to recruit and may struggle to complete the spirometry test due to physical limitations~\cite{10.1145/3570167}. In the future, we aim to incorporate variable labeling costs to make our systems more practical.
    
    \item Currently, we do not consider a stopping criteria in active learning which can lead to diminishing returns or worsened performance with additional samples. Future work will include heuristics for stopping when performance plateaus, such as not adding a pool point if its distance to existing training points is below a threshold, or using confidence based stopping criteria~\cite{zhu2010confidence} to stop when the confidence on the unlabeled data exceeds a threshold value.
    
    \item In our current analysis, we add a single pool point to the training dataset. In the future, we plan to also allow adding multiple pool points in one go. This translates to calling multiple participants in a single session for collecting ground truth making the setting more practical. Prior work~\cite{kirsch2019batchbald} demonstrates how mutual information between a batch of points and model parameters can be used to select multiple informative points for active learning. 
    
    \item Generation of features for every participant is a tedious task. We plan to explore neural networks like 1D CNNs that use audio signals as inputs and provide us with the FVC and FEV1 values.
\end{itemize}

\section{Conclusion}
\noindent Collecting ground truth for medical applications is challenging and incurs substantial costs. Intelligent querying techniques are essential to address the issue of limited datasets in this domain. This study presented active learning as a structured approach for strategically recruiting new participants from a pool, whose ground truth help improve overall model performance. We introduced various active learning and modeling strategies designed to efficiently acquire data by considering multiple health indices. Our proposed active learning methods outperformed random acquisition and adhered to the error thresholds set by ATS. 

\noindent We explored both single-task and multi-task active learning settings, achieving errors of approximately $\sim$3\% in both settings. While multi-task active learning yielded higher intermediate errors as compared to single task, it proved to be a more practical solution. Multi-task active learning achieved errors as low as 3.3\% with $\sim$20 participants, potentially saving approximately 40 hours of data collection time. By leveraging this approach, we significantly reduce the burden of ground truth collection, making wearable spirometry more accessible for future medical research.

\bibliographystyle{IEEEtran}  
\bibliography{main.bib}

\end{document}